\documentclass[conference]{IEEEtran}
\ifCLASSINFOpdf
  \usepackage[pdftex]{graphicx}
\else
\fi
\begin{document}
%
\title{Abnormality Detection in Mammography using Deep Convolutional Neural Networks}

\author{\IEEEauthorblockN{Pengcheng Xi}
\IEEEauthorblockA{Carleton University and\\National Research Council Canada\\
Ottawa, Ontario, Canada\\
pengcheng.xi@nrc-cnrc.gc.ca}
\and
\IEEEauthorblockN{Chang Shu}
\IEEEauthorblockA{Computer Vision and Graphics\\
National Research Council Canada\\
Ottawa, Ontario, Canada\\
chang.shu@nrc-cnrc.gc.ca}
\and
\IEEEauthorblockN{Rafik Goubran}
\IEEEauthorblockA{Systems and Computer Engineering\\
Carleton University\\
Ottawa, Ontario, Canada\\
goubran@sce.carleton.ca}
}

%


\maketitle

\begin{abstract}
Breast cancer is the most common cancer in women worldwide. The most common screening technology is mammography. To reduce the cost and workload of radiologists, we propose a computer aided detection approach for classifying and localizing calcifications and masses in mammogram images. To improve on conventional approaches, we apply deep convolutional neural networks (CNN) for automatic feature learning and classifier building. In computer-aided mammography, deep CNN classifiers cannot be trained directly on full mammogram images because of the loss of image details from resizing at input layers. Instead, our classifiers are trained on labelled image patches and then adapted to work on full mammogram images for localizing the abnormalities. State-of-the-art deep convolutional neural networks are compared on their performance of classifying the abnormalities. Experimental results indicate that VGGNet receives the best overall accuracy at 92.53\% in classifications. For localizing abnormalities, ResNet is selected for computing class activation maps because it is ready to be deployed without structural change or further training. Our approach demonstrates that deep convolutional neural network classifiers have remarkable localization capabilities despite no supervision on the location of abnormalities is provided.

\end{abstract}


%
\IEEEpeerreviewmaketitle

\section{Introduction}

According to World Health Organization (WHO), breast cancer is the most common cancer in women both in the developed and the developing world \cite{WHO2011}. Moreover, there is an increasing incidence of breast cancer in the developing world because of the increase in life expectancy, urbanization and adoption of western lifestyles. Although some risk reduction can be achieved with prevention, early detection for improving breast cancer outcome and survival remains the cornerstone of breast cancer control \cite{WHO2011}.

Mammography is the most common breast screening technology. There are several imaging techniques for examining the breast, including ultrasound, magnetic resonance imaging (MRI), X-ray imaging and emerging technologies such as molecular breast imaging and digital breast tomosynthesis (DBT). Mammography is a type of imaging that uses a low-dose X-ray system to examine the breast and is the most reliable method for screening breast abnormalities \cite{advance2009} before they become clinically palpable.

There are two types of examinations in mammography: screening and diagnostic. Screening mammography is for detecting breast cancer in an asymptomatic population while diagnostic mammography is a follow-up exam on patients who have already demonstrated abnormal clinical findings \cite{advance2009}. Screening mammography generally consists of four views, with two views of each breast: the craniocaudal (CC) view and the mediolateral oblique (MLO) view. Besides the two views, additional diagnostic mammography may offer in-depth look at suspicious areas.

One of the challenges in mammography is low contrast in mammogram images. This poses difficulties for radiologists to interpret results. Double reading of mammograms has been advocated to lower the rate of false positives and negatives \cite{doubleReading}; however, the cost and workload associated with double reading are high. Therefore, computer aided detection (CADe) and computer aided diagnosis (CADx) of abnormalities in mammography have been introduced. While CADx has not been approved for clinical use, CADe is playing an increasingly important role in breast cancer screening \cite{advance2009} \cite{survey2016}.

Computer aided detection is a pattern recognition process that aids radiologists in detecting potential abnormalities such as calcifications, masses, and architectural distortions \cite{cad05}. It identifies suspicious features in the radiology images and brings them to the attention of radiologists \cite{cad05}. In its current use, the radiologists first review the exam, activates the CAD software and then re-evaluates the CAD-marked areas of concern before writing the report \cite{cad05}.

Because of the medical significance of screening breast cancer, there has been considerable effort on developing CAD approaches for detecting abnormalities, including calcifications, masses, architectural distortion and bilateral asymmetry \cite{advance2009} \cite{survey2016} \cite{JALALIAN2013420} \cite{7244993}. Traditional CAD approaches rely on manually designed image features \cite{advance2009} \cite{survey2016} in detecting subtle yet crucial abnormalities in mammograms. In general, the detection of calcifications followed the procedure of image enhancement, stochastic modelling, frequency decomposition and machine learning; the detection of masses have relied on pixel-based and region-based approaches \cite{advance2009}.

Recent advances in deep neural networks have enabled automatic feature learning from large amount of training data, providing an end-to-end solution from feature extraction to classifier building \cite{alexnet} \cite{DBLP:journals/corr/SimonyanZ14a} \cite{googlenet} \cite{DBLP:journals/corr/HeZRS15}. Moreover, this learning scheme is robust to dataset noise, making it suitable for detecting abnormalities in mammography.

In this work, we present an abnormality detection approach using deep Convolutional Neural Networks (CNN). Using transfer learning \cite{murmur}, we fine tune pre-trained deep CNNs on cropped image patches of calcifications and masses. After feeding a full mammogram image to input of the CNN tuned on patch images, we compute Class Activation Maps (CAM) for localizing abnormalities \cite{DBLP:journals/corr/HeZRS15}.

Our contributions are three-fold:
\begin{itemize}
\item Significantly leveraged deep CNNs' hierarchical feature extraction capabilities through transfer learning. This enables automatic extraction of features for classifying and localizing calcification and mass in mammograms.
\item Compared the performance of state-of-the-art deep CNN architectures by training with a limited dataset without over-fitting.
\item Successfully adapted patch-based CNN classifiers to full mammogram images for the localization of abnormalities without segmentation.
\end{itemize}

\section{Literature Review}



We review computer-aided approaches to detecting and classifying the two main abnormalities found in screening mammography: micro-calcification (MC) and mass. Most approaches to detecting calcifications follow a similar procedure: image enhancement, segmentation or extracting Region of Interests (ROIs), feature computation and classification. Mass detection algorithms first detect suspicious regions in a mammogram and then classify it as mass or normal tissues.

MCs are tiny deposits of calcium that appear as bright spots in mammograms. Filter banks were used to decompose mammogram images followed by ROI selection and Bayesian classifications \cite{1580833} \cite{4381127}. Pal et al. \cite{PAL20082625} introduced a multi-stage system for detecting MCs in mammograms. They used a back-propagation neural network to find candidate calcification regions first, cleaned network output to remove thin elongated structures and used a measure of local density for final classification. Similarly, Harirchi et al. \cite{5597590} applied a two-level algorithm for the detection of MCs using diverse-Adaboost-SVM. Six features (four wavelet plus two gray level features) were computed for neural network to detect candidate MC pixels. As a result, 25 features from candidate MCs were extracted and further reduced with geometric linear discriminant analysis (GLDA). The classifier was built with diverse Adaboost SVM. Oliver et al. \cite{Oliver:2012:AMC:2125394.2125474} extracted local features for morphology of MCs and then used a learning approach to select the most salient feature for a boosted classifier. Zhang et al. \cite{ZHANG20121062} enhanced the MCs using well-designed filters and then conducted subspace learning for feature selection. A twin SVM (TWSVM) was used for classification.

A mass in mammogram is defined as a space-occupying lesion seen in more than one projection \cite{acr}. The general procedure for detecting mass is first to detect suspicious regions, then extract shape and texture features, and finally detect mass regions through classification or removing false positive regions \cite{survey2016}. Petrosian et al. \cite{Petrosian} used texture features to distinguish mass and non-mass regions. Petrick et al. \cite{481441} used an adaptive density-weighted contrast enhancement filter to obtain potential masses and used Laplacian Gaussian for edge detection. Morphological features were extracted for classifying normal and mass ROIs. Cascio et al. \cite{1710274} first segmented the boundary of ROI using an edge-based approach and then computed geometric and shape features. Neural networks were trained to distinguish true mass from normal regions.

While previous classifiers mostly used shallow neural networks, recent years witnessed great advancement on applying deep learning to computer aided detection. Wang et al. \cite{DBLP:journals/corr/WangPLLBS17} introduced ChestX-ray8, a hospital-scale chest X-ray database, and provided benchmarks on weakly-supervised classification and localization of common thorax diseases. They applied deep CNNs and added transition layers to produce heatmap for localization. Following this work, Rajpurkar et al. \cite{DBLP:journals/corr/abs-1711-05225} introduced CheXNet, a 121-layer Dense Convolutional Network (DenseNet) trained on the ChestX-ray 14 dataset, producing radiologist-level pneumonia detection. Moreover, Rajpurkar et al. \cite{mura} introduced MURA dataset for detecting radiologist-level abnormality in musculo-skeletal radiographs.

Machine learning has also been widely applied to medical measurements and imaging applications. Rosati et al. \cite{7533734} used multiparametric MRI along with a clustering procedure based on self-organizing map (SOM) to improve the detection of prostate cancer. Andria et al. \cite{7533723} investigated the relation between the radiation dose on patient and the resulting image quality, through comparing the tomosynthesis performance with 2D digital mammography. Roza et al. \cite{7985908} presented an artificial neural network (ANN) and feature extraction methods to identify two types of arrhythmias in ECG signals. Alkabawi et al. \cite{7985847} proposed an approach for computer-aided classification of multi-types of dementia using convolutional neural networks. The proposed approach outperforms the state-of-the-art CAD methods.

\section{Methodology}

Computer-aided mammography is a challenging problem and cannot be treated as an image classification task. The reason is that abnormalities within a whole image are located in small regions. For example, a typical full mammogram with a resolution of 3000x4600 (width and height in pixels) contains an abnormality region of size only about 200x200 (pixels). Training recent deep CNNs requires resizing full images to 224x224 (pixels) at input layer, making it difficult to train and detect abnormalities. To deal with this challenge, we propose training deep CNNs on cropped image patches (labelled ROIs) and adapting them to full mammogram images.

\begin{figure}[htbp]
\centerline{\includegraphics[width=1.0\linewidth]{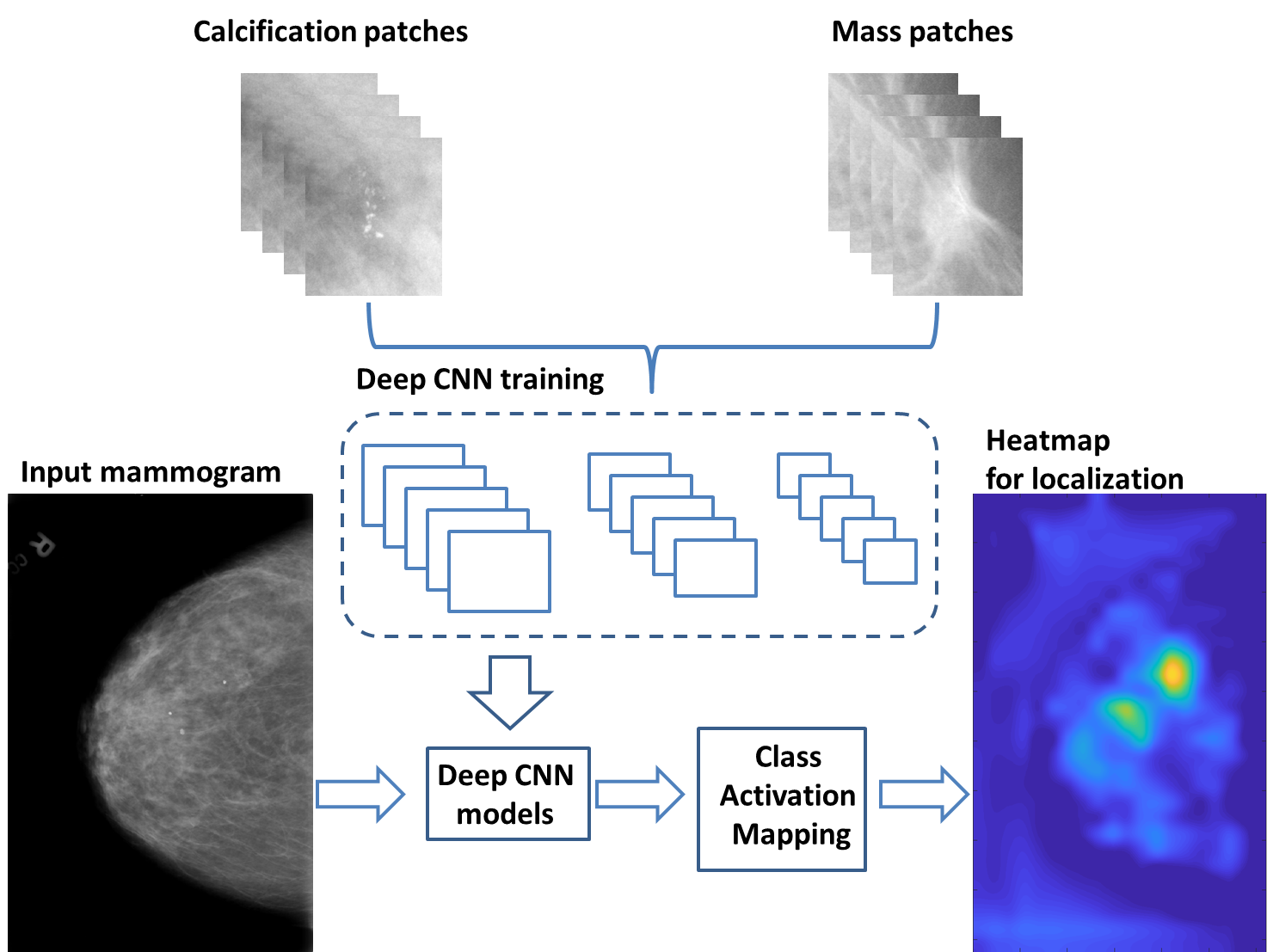}}
\caption{Diagram of our approach.}
\label{approach}
\end{figure}

Figure \ref{approach} illustrates the data-flow of our approach. With training image patches from calcification and mass cases, a binary classifier is trained with state-of-the-art deep CNN architectures using transfer learning \cite{murmur}. The pre-trained CNNs are modified at output layers to have two output classes.  The output layers are then fine-tuned while the first part of the network is frozen.

The fine-tuned patch neural network is then used to localize mammographic abnormalities in full-size mammograms. Traditional approaches used the classifier to scan the whole image with a sliding window and therefore have a low efficiency \cite{murmur}. In contrast, our approach enables localizing abnormalities in one single forward pass. Feeding the full-size mammogram image into the patch classifier and computing class activation mapping \cite{zhou2015cnnlocalization} near the end of the output layers produces a heatmap for the localization of abnormalities. The computation of CAM is explained in more details at section \ref{cam}.

\subsection{Data Selection}

\begin{figure}[htbp]
\centerline{\includegraphics[width=1.0\linewidth]{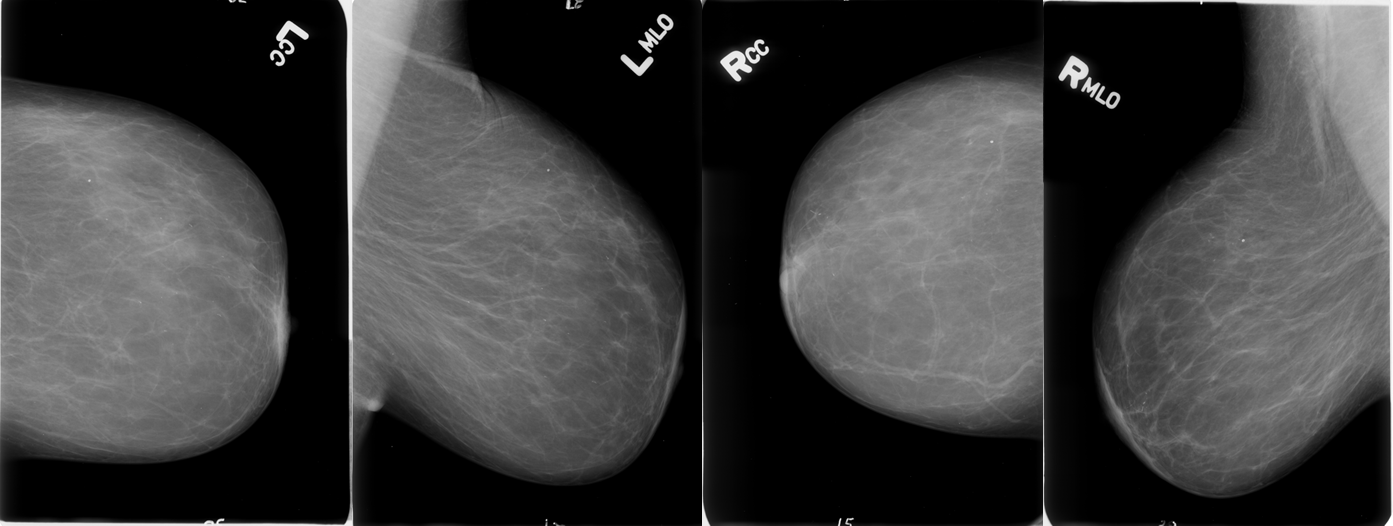}}
\caption{Mammograms of a patient in DDSM dataset (different views from left to right: left CC, left MLO, right CC and right MLO).}
\label{mammograms}
\end{figure}

In mammography, there is a lack of standard evaluation data and most CAD algorithms are evaluated on private dataset. Most mammographic databases are not publicly available. This poses a challenge to compare performance of methods or to replicate prior results. The most commonly used databases are the Mammographic Image Analysis Society (MIAS) database \cite{MIAS} and the Digital Database for Screening Mammography (DDSM) \cite{DDSM}. MIAS contains left and right breast images for 161 patients. There are 208 normal, 63 benign and 51 malignant images. It also includes radiologist's `truth'-markings on the locations of any abnormalities that may be present. DDSM is the largest mammography dataset that is publicly available. The database contains approximately 2,500 studies, each includes two images of each breast, along with associated patient information and image information. Images containing suspicious areas have associated pixel-level ``ground truth" about the locations and types of suspicious regions. Sample mammograms of a patient are shown in Figure \ref{mammograms}.


\begin{figure}[htbp]
\centerline{\includegraphics[width=1.0\linewidth]{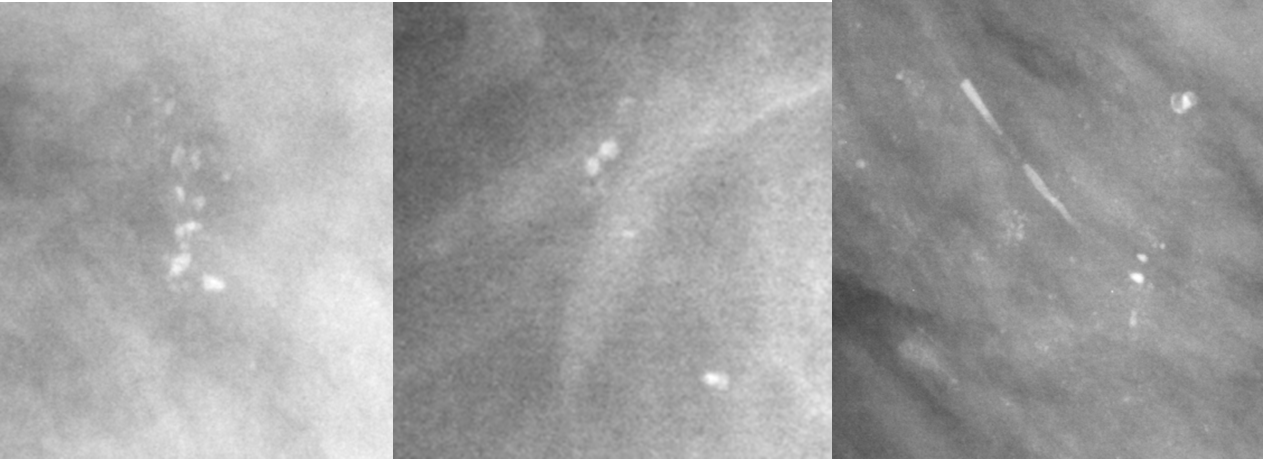}}
\caption{Sample images of calcification patches in CBIS-DDSM.}
\label{calcPat}
\end{figure}

\begin{figure}[htbp]
\centerline{\includegraphics[width=1.0\linewidth]{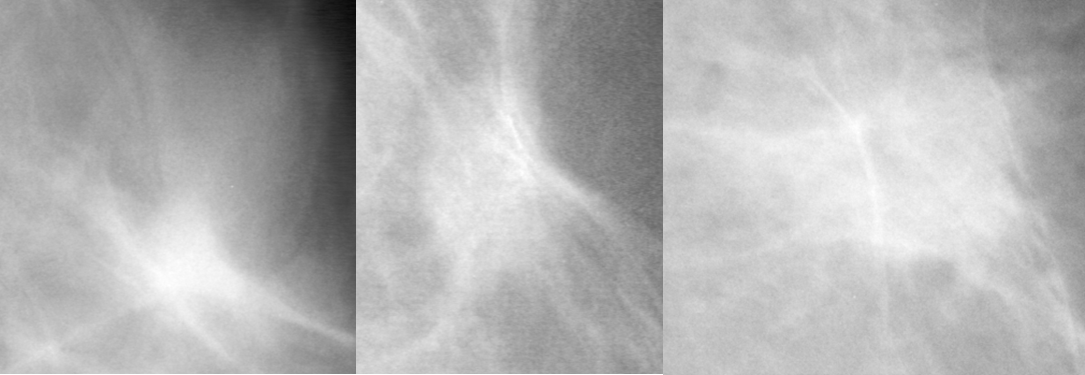}}
\caption{Sample images of mass patches in CBIS-DDSM.}
\label{massPat}
\end{figure}

Recently, Lee et al. \cite{CBIS-DDSM} released an updated and standardized version of the DDSM for the evaluation of CAD systems in mammography. Their dataset, the CBIS-DDMS (Curated Breast Imaging Subset of DDSM), includes decompressed images, data selection and curation by trained mammographers, updated mass segmentation and bounding boxes, and pathologic diagnosis for training data. The dataset contains 753 calcification cases and 891 mass cases. Sample image patches are shown in Figure \ref{calcPat} and \ref{massPat}.

We use image patches from CBIS-DDSM for classification and test on full mammograms for localization. We merge the training and testing dataset in CBIS-DDSM and conduct new 85/15 split for training and testing sets. The number of image patches are listed in Table \ref{dataTable}.

\begin{table}[htbp]
\caption{Size of training and testing image patches for CBIS-DDSM.}
\begin{center}
\begin{tabular}{|l||c|c|c|}
\hline
\textbf{Abnormality} & \textbf{Training} & \textbf{Testing} & \textbf{Overall} \\
\hline\hline
Calcification   & 1284 & 227 & 1511 \\
Mass    & 1353 & 239 & 1592 \\
\hline
\end{tabular}
\label{dataTable}
\end{center}
\end{table}


\subsection{Data Augmentation}

To avoid over-fitting during training, we applied the following data augmentation on the training data: random rotation between zero and 360 degrees, random X and Y reflections. It is based on our observation of the variations within the training and testing dataset.


\subsection{Architectures of Deep CNN}

In visual computing, tremendous progress has been made in object classification and recognition thanks to the availability of large scale annotated datasets such as ImageNet Large Scale Visual Recognition Competition (ILSVRC) \cite{ILSVRC15}. The ImageNet dataset contains over 15 million annotated images from a total of over $22,000$ categories.

Recent years witnessed great performance advancement on ILSVRC using deep CNNs. Comparing to traditional hand-crafted image features, deep CNNs automatically extract features from a large dataset for tasks they are trained for. In this work, we adapt four of the best-performing models in recent ImageNet challenges and compare their performance on classifying calcification and mass in mammograms.

\begin{itemize}
\item AlexNet. In 2012, Krizhevsky et al. \cite{alexnet} entered ImageNet ILSVRC with a deep CNN and achieved top-5 test error rate of $15.3\%$, compared to $26.2\%$ achieved by the second-best entry. The network was made up of 5 conv layers, max-pooling layers, dropout layers, and 3 fully connected layers. This work led to a series of deep CNN variants in the following years which consistently improved the state-of-the-art in the benchmark tasks. 
\item VGGNet. In 2014, Simonyan and Zisserman \cite{DBLP:journals/corr/SimonyanZ14a} introduced a deeper 19-layer CNN and achieved top result in the localization task of ImageNet ILSVRC. The network used very small 3x3 convolutional filters and showed significant improvement. This influential work indicated that CNNs need to have a deep network of layers in order for the hierarchical feature representations to work.
\item GoogLeNet. In 2014, Szegedy et al. \cite{googlenet} introduced a deeper CNN to ILSVRC and achieved top 5 error rate of $6.7\%$. Instead of sequentially stacking layers, this network was one of the first CNNs that used parallel structures in its architecture (9 Inception modules with over 100 layers in total).
\item ResNet. In 2015, He et al. \cite{DBLP:journals/corr/HeZRS15} introduced a new 152-layer network architecture and set new records in ILSVRC. ResNet achieved $3.57\%$ error rate in the classification task. The residual learning framework is 8 times deeper than VGGNet but still has lower complexity.
\end{itemize}

All the deep CNN architectures were designed for a 1000-class classification task. To adapt them to our task, the last three layers were removed from each network. Three new layers (fully connected layer, soft-max layer and classification layer) were appended to the remaining structure of each network. Higher learning rates were set for the newly added fully connected layers so that the first part of each network remains relatively unchanged during training and the newly added layers get fine-tuned on our dataset. Five-fold cross validation is used to train and test the robustness of each architecture. 

\subsection{Class Activation Maps} \label{cam}

Class Activation Mapping (CAM) is a technique for identifying regions in an image using a CNN for a specific class \cite{zhou2015cnnlocalization}. In other words, CAM identifies image regions relevant to a class. It allows re-using classifiers for localization purpose, even when no training data on locations are available. It also demonstrates that CNNs have a built-in attention capability.

\begin{figure}[htbp]
\centerline{\includegraphics[width=1.0\linewidth]{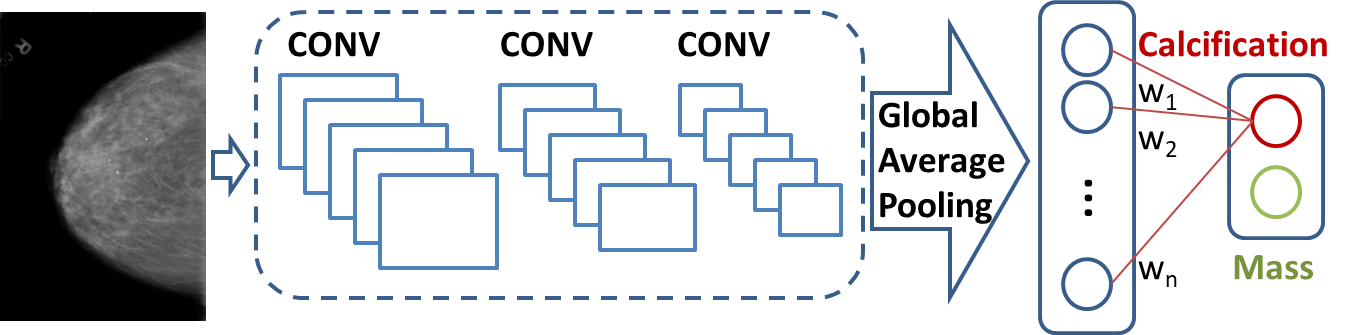}}
\caption{Class activation mapping for heatmap production.}
\label{CAM}
\end{figure}

Computing CAM for mammograms is explained in Figure \ref{CAM}. A deep CNN needs to be cut after the last convolution layer and a global average pooling layer and a fully connected layer are appended. The new model needs to be retrained for learning the weights $w_i$ ($i=1,2,...n$) at the output layer. Within the four selected deep CNN architectures, ResNet already has the required architecture and is therefore selected for computing CAM.

A full mammogram is fed into the fine-tuned patch classifier using ResNet. The feature maps from the output of the last convolutional layer are denoted as $f_i$ ($i=1,2,...n$). We can identify the importance of the image regions by projecting back the weights of the output layer onto the convolutional feature maps \cite{zhou2015cnnlocalization} through:

\begin{equation}
CAM = \sum_{i=1}^{n}{w_i f_i}
\end{equation}

The output CAM is then displayed for visualization and verification.

\section{Results}

\subsection{Comparison of Different Deep CNN Architectures}
We set the following parameters for training each modified deep CNN: Stochastic Gradient Descent with Momentum (SGDM) as the optimization algorithm, batch size of 16, initial learning rate as $1e-4$, and the learning rate factor for the last fully connected layer as 20.0. Each network stops from further training if the mean accuracy on the fifty most recent batches reaches $99.5\%$ or if the number of epochs reaches maximum setting of 200. All the models are trained on a workstation with an NVIDIA GeForce GTX TITAN X GPU (one hour for AlexNet, eight hours for VGGNet, two hours for GoogLeNet, and four hours for ResNet). The final size of fine-tuned VGGNet is about 20 times that of GoogLeNet, with in-between sizes for AlexNet and ResNet.

\begin{table}[htbp]
\caption{Mean Classification Accuracies of Deep CNNs}
\begin{center}
\begin{tabular}{|l||c|c|c|}
\hline
\textbf{Model} & \textbf{Calcification} & \textbf{Mass} & \textbf{Overall Accuracy} \\
\hline\hline
AlexNet & 88.81\% & 93.64\% & 91.23\% \\ 
VGGNet & \textbf{92.42\%} & 92.64\% & \textbf{92.53\%} \\ 
GoogleNet & 87.14\% & \textbf{95.06\%} & 91.10\% \\ 
ResNet & 90.22\% & 93.39\% & 91.80\% \\ 
\hline
\end{tabular}
\label{result}
\end{center}
\end{table}

Running cross-validation on training and testing datasets and computing mean accuracies across the five folds give the final accuracy results in Table \ref{result}. VGGNet achieves the highest accuracy for classifying calcifications and GoogleNet receives the best performance for classifying masses. The highest overall accuracy is also achieved by VGGNet at 92.53\%.

\subsection{Localization Results using Class Activation Mapping}

\begin{figure}[htbp]
\centerline{\includegraphics[width=1.0\linewidth]{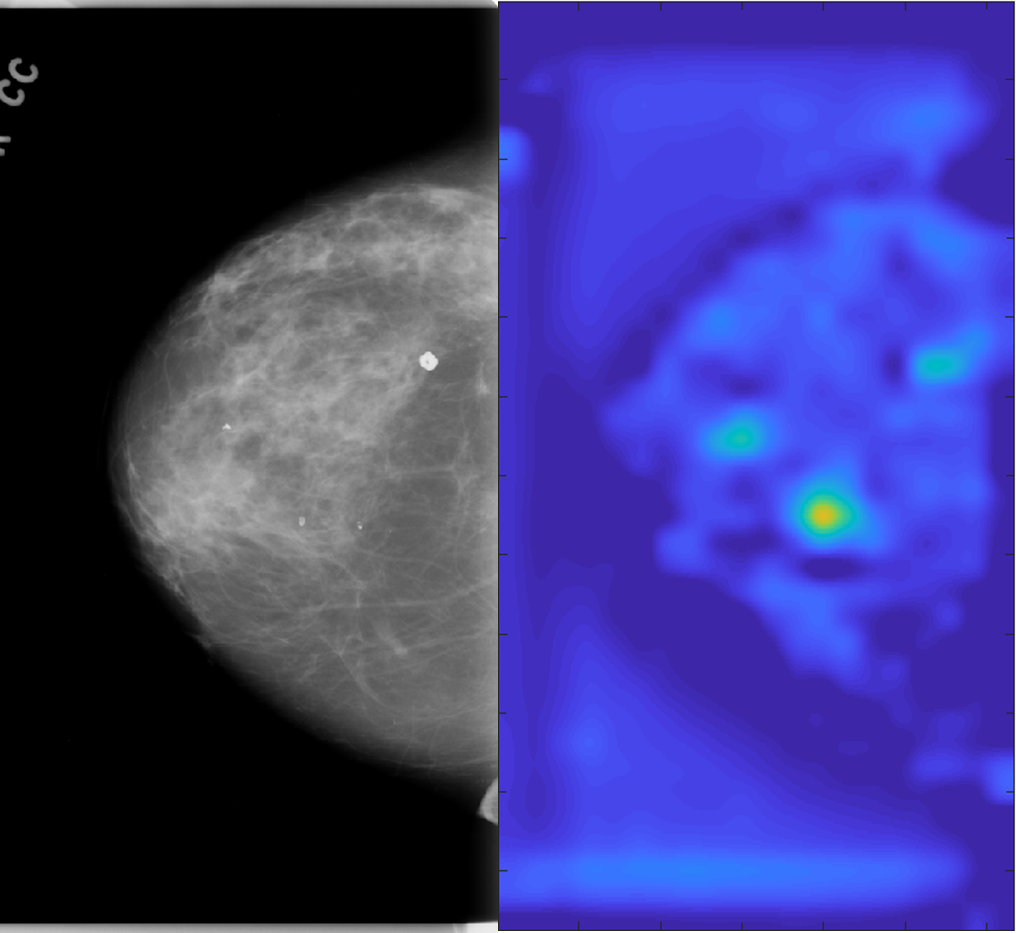}}
\caption{Calcification localization result (left: full mammogram, right: class activation map output.}
\label{calcLoc}
\end{figure}

We use the fine-tuned ResNet to compute class activation mapping for localizing abnormalities. The selection is based on the fact that ResNet is ready to be used for computing CAM without further training. Without losing generality, we use one full mammogram image from the calcification class, feed it to ResNet and compute the CAM. The result is shown in Figure \ref{calcLoc}. The heatmap on the right highlights the location of calcifications found from the input mammogram. The highlighted regions correspond to the calcifications within the full mammogram (best viewed in color).

\begin{figure}[htbp]
\centerline{\includegraphics[width=1.0\linewidth]{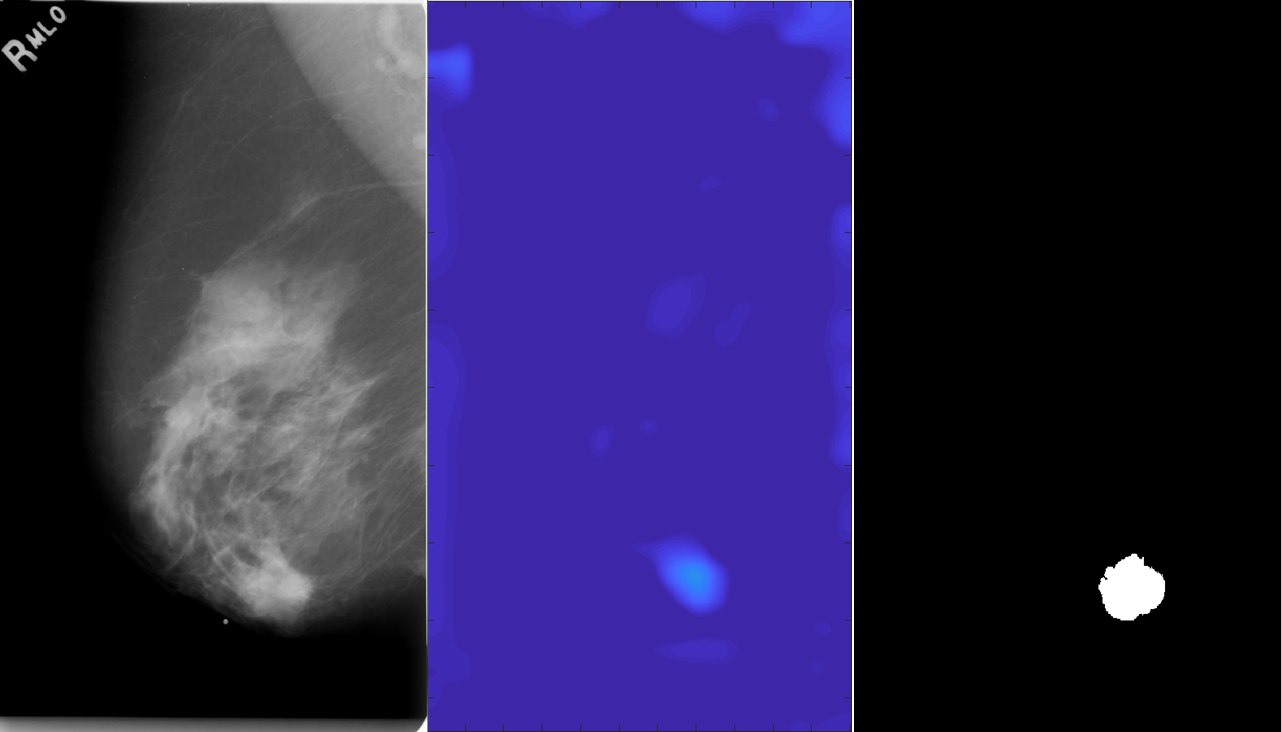}}
\caption{Mass localization result (from left to right: full mammogram, class activation map output, ground-truth binary mask image).}
\label{massLoc}
\end{figure}

Similarly a full mammogram from the mass class is fed into the ResNet for computing the CAM. Results are demonstrated in Figure \ref{massLoc}. To add to the comparison, we also include the ground-truth binary mask image provided by the training dataset. The highlighted heatmap region corresponds to the identified abnormality region labelled in the binary mask image.

\section{Discussions}

Because of the low-contrast and noise in mammogram images, it is challenging to train classifiers on calcification and mass cases. Deep neural networks has a limitation on the size of input images (224x224 or 227x227 in pixels). Resizing mammogram images to these sizes will inevitably reduce the quality of images and may also lose the subtle details that are needed for classification. Therefore we propose training classifiers from cropped batch images in order to catch the difference between calcification and mass cases, and apply the trained deep CNN models onto full-size mammogram images. Using a technique called class activation mapping, we successfully reuse the patch classifier for the localization of abnormalities in full mammogram images.

\section{Conclusion}

We successfully apply deep convolutional neural networks to localizing calcifications and masses in mammogram images without training directly on the full images. This is achieved by conducting the training on cropped image patches through transfer learning and data augmentation. State-of-the-art deep CNN architectures are trained and compared on their performance of classifying the abnormalities. Moreover, we successfully adapt the patch classifier to localizing abnormalities in full mammogram images through class activation mapping.

At the time of preparing this paper, we have found no publications on using CBIS-DDSM; therefore, our results provide a baseline for future studies on improving the performance of detecting calcification and mass in computer-aided mammography. Our future work includes extending the approach to computer aided diagnosis (benign or malignant) using mammograms.






%

%
%

{\small
\bibliographystyle{IEEEtranS}
\bibliography{bare_conf}
}

\end{document}